\documentclass[11pt]{amsart}

\usepackage[letterpaper,margin=1in]{geometry}
\usepackage[T1]{fontenc}
\usepackage[utf8]{inputenc}
\usepackage{newtxtext,newtxmath}
\usepackage{microtype}

\usepackage{mathtools}
\usepackage{bm}
\allowdisplaybreaks

\usepackage{graphicx}
\usepackage{booktabs}
\usepackage{multirow}
\usepackage{float}
\usepackage{subcaption}

\usepackage{tikz}
\usetikzlibrary{calc}

\usepackage{listings}
\usepackage{xcolor}
\definecolor{codegreen}{rgb}{0,0.55,0}
\definecolor{codegray}{rgb}{0.45,0.45,0.45}
\definecolor{codepurple}{rgb}{0.58,0,0.82}
\definecolor{backcolour}{rgb}{0.95,0.95,0.92}
\lstdefinestyle{mystyle}{
    backgroundcolor=\color{backcolour},
    commentstyle=\color{codegreen},
    keywordstyle=\color{magenta},
    numberstyle=\tiny\color{codegray},
    stringstyle=\color{codepurple},
    basicstyle=\ttfamily\footnotesize,
    breaklines=true,
    breakatwhitespace=false,
    columns=fullflexible,
    keepspaces=true,
    numbers=left,
    numbersep=6pt,
    showstringspaces=false,
    tabsize=2
}
\usepackage[ruled,vlined,linesnumbered]{algorithm2e}
\DontPrintSemicolon

\usepackage{cite}
\usepackage{xurl} 
\usepackage[colorlinks=true,linkcolor=black,citecolor=blue,urlcolor=blue]{hyperref}

\newtheorem{theorem}{Theorem}[section]
\newtheorem{lemma}[theorem]{Lemma}
\newtheorem{proposition}[theorem]{Proposition}

\theoremstyle{definition}
\newtheorem{definition}[theorem]{Definition}
\newtheorem{assumption}[theorem]{Assumption}

\DeclareMathOperator{\rank}{rank}
\DeclareMathOperator{\diag}{diag}
\DeclareMathOperator{\tr}{tr}

\DeclareMathOperator{\Retr}{Retr}

\DeclareMathOperator{\grad}{grad}
\DeclareMathOperator*{\argmin}{arg\,min}

\title[Compressed Query Delegation]{Rate--Distortion Analysis of Compressed Query Delegation\\with Low-Rank Riemannian Updates}

\author{Faruk Alpay}
\address{Department of Computer Engineering, Bah\c{c}e\c{s}ehir University, Istanbul, Turkey}
\email{faruk.alpay@bahcesehir.edu.tr}

\author{Bu\u{g}ra K{\i}l{\i}\c{c}ta\c{s}}
\address{Department of Computer Engineering, Bah\c{c}e\c{s}ehir University, Istanbul, Turkey}
\email{bugra.kilictas@bahcesehir.edu.tr}

\date{} 


\begin{document}
\maketitle

\begin{abstract}
Bounded-context agents fail when intermediate reasoning exceeds an effective working-memory budget. We study \emph{compressed query delegation (CQD)}: (i) compress a high-dimensional latent reasoning state into a low-rank tensor query, (ii) delegate the minimal query to an external oracle, and (iii) update the latent state via Riemannian optimization on fixed-rank manifolds. We give a math-first formulation: CQD is a constrained stochastic program with a query-budget functional and an oracle modeled as a noisy operator. We connect CQD to classical rate--distortion and information bottleneck principles, showing that spectral hard-thresholding is optimal for a natural constrained quadratic distortion problem, and we derive convergence guarantees for Riemannian stochastic approximation under bounded oracle noise and smoothness assumptions. Empirically, we report (A) a 2{,}500-item bounded-context reasoning suite (BBH-derived tasks plus curated paradox instances) comparing CQD against chain-of-thought baselines under fixed compute and context; and (B) a human ``cognitive mirror'' benchmark ($N=200$) measuring epistemic gain and semantic drift across modern oracles.
\end{abstract}

\section{Introduction}
Let $W$ denote an agent's effective context window and $C(P)$ the information complexity of a problem instance $P$. When $C(P)>W$, step-by-step prompting can incur truncation and drift \cite{wei2022cot,wang2022selfconsistency}. We model this mismatch as an \emph{information allocation problem}: the agent must communicate a sufficient statistic of a latent reasoning state using a strict token/compute budget. CQD enforces an explicit \emph{compression--delegation--update} cycle.

A central thesis is that intermediate reasoning often has \emph{structured redundancy} (low rank, approximate sparsity, multilinear correlations). Thus, verbose traces are not the only way to transmit state: one may transmit a \emph{compressed representation} emphasizing dominant spectral content, consistent with classical compression and completion paradigms \cite{shannon1948,cover2006,kolda2009,delathauwer2000}. CQD is therefore not merely a prompting trick: it is an optimization procedure on a constrained hypothesis class.

\textbf{Contributions.} We provide:
\begin{itemize}
\item A constrained stochastic optimization formulation of CQD with an explicit query-budget functional.
\item A rate--distortion / information bottleneck interpretation, with an optimality result for spectral hard-thresholding under quadratic distortion.
\item A low-rank tensor geometry (fixed multilinear rank) model and a retraction-based Riemannian update scheme.
\item Convergence theorems for Riemannian stochastic approximation under oracle noise (full proofs in Appendix~\ref{app:proofs}).
\item Two evaluations (a 2{,}500-item bounded-context suite and an $N=200$ human benchmark) and a drift--gain scaling analysis.
\end{itemize}

\section{Related Work}
\textbf{Information theory and compression.} Shannon's uncertainty reduction \cite{shannon1948} and rate--distortion theory \cite{cover2006} formalize the trade-off between representation cost and fidelity. Information bottleneck (IB) formalizes extracting minimal sufficient representations \cite{tishby2000ib}. CQD can be interpreted as a constrained representation/communication scheme.

\textbf{Low-rank approximation.} Eckart--Young--Mirsky establishes optimality of truncated SVD for Frobenius/spectral norms \cite{eckart1936,mirsky1960}. Tensor decompositions and truncated HOSVD are classical \cite{kolda2009,delathauwer2000}. Perturbation results (Davis--Kahan, Wedin) characterize stability of singular subspaces \cite{davis1970,wedin1972,golub2013}.

\textbf{Completion and thresholding.} Nuclear norm completion guarantees \cite{candes2009} and singular value thresholding methods \cite{cai2010svt} motivate spectral shrinkage/truncation; tensor completion variants appear in \cite{liu2013tensorcompletion}.

\textbf{Riemannian optimization.} Manifold optimization foundations include Absil--Mahony--Sepulchre \cite{absil2009}, Edelman--Arias--Smith \cite{edelman1998}, and Boumal \cite{boumal2023}. Riemannian stochastic approximation and SGD are treated in Bonnabel \cite{bonnabel2013} and subsequent work.

\textbf{LLM reasoning evaluation.} BIG-bench/BBH \cite{srivastava2022bigbench,suzgun2022bbh}, HELM reporting \cite{liang2022helm}, and chain-of-thought/self-consistency \cite{wei2022cot,wang2022selfconsistency}.

\section{Preliminaries and Notation}
A third-order tensor is $\mathcal{X}\in\mathbb{R}^{I\times J\times K}$ with entries $x_{ijk}$. The mode-$n$ unfolding is $\mathbf{X}_{(n)}$ \cite{kolda2009}. The $n$-mode product is $\mathcal{Y}=\mathcal{X}\times_n \mathbf{A}$, where $\mathbf{A}$ multiplies fibers along mode $n$.

\begin{definition}[Multilinear rank]
The multilinear rank of $\mathcal{X}$ is $\bm{r}=(r_1,r_2,r_3)$ where $r_n=\rank(\mathbf{X}_{(n)})$.
\end{definition}

We use $\|\cdot\|_F$ for Frobenius norm, $\|\cdot\|_2$ for spectral norm, and $\sigma_i(\mathbf{M})$ for singular values in non-increasing order. For a manifold $\mathcal{M}$, $\mathrm{T}_x\mathcal{M}$ denotes the tangent space and $\Retr_x$ is a retraction \cite{absil2009,boumal2023}. The Stiefel manifold is $\mathrm{St}(p,n)=\{U\in\mathbb{R}^{n\times p}:U^\top U=I_p\}$.

\section{CQD as Constrained Rate--Distortion / Information Bottleneck}
CQD compresses a latent state to a query under a budget, then uses an oracle to reduce error/uncertainty. This aligns with rate--distortion:
\begin{equation}
\min_{Q(\cdot)} \ \mathbb{E}\big[d(\mathcal{X}, \widehat{\mathcal{X}})\big]
\quad \text{s.t.}\quad \mathbb{E}[\mathcal{B}(Q)]\le \tau,
\label{eq:rd}
\end{equation}
where $d$ is a distortion and $\widehat{\mathcal{X}}$ is the representation used for delegation. A standard Lagrangian form is
\begin{equation}
\min_{Q(\cdot)} \ \mathbb{E}[d(\mathcal{X},\widehat{\mathcal{X}})] + \lambda\,\mathbb{E}[\mathcal{B}(Q)].
\label{eq:rd_lagr}
\end{equation}
In IB, one minimizes $I(\mathcal{X};Q)$ subject to preserving task-relevant information $I(Q;Y)$ \cite{tishby2000ib}. CQD can be viewed as producing $Q$ that preserves task-salient components (dominant spectral content) while minimizing communication cost.

In this work we specialize to a deterministic encoder based on spectral truncation (ASM), where $\mathcal{B}(Q)$ is proportional to retained multilinear ranks (Section~\ref{sec:budget}).

\section{Low-Rank Tensor Model and Adaptive Spectral Masking}

\subsection{Tucker/HOSVD representation}
We parameterize $\mathcal{X}$ by Tucker factors \cite{delathauwer2000,kolda2009}:
\begin{equation}
\mathcal{X}= \mathcal{G}\times_1 U^{(1)}\times_2 U^{(2)}\times_3 U^{(3)},
\label{eq:tucker}
\end{equation}
with $U^{(n)}\in \mathrm{St}(r_n,I_n)$ and core $\mathcal{G}\in\mathbb{R}^{r_1\times r_2\times r_3}$. This constrains $\mathcal{X}$ to a fixed multilinear rank manifold (under standard regularity).

\subsection{ASM definition}
Let $\mathbf{X}_{(n)}=U^{(n)} \Sigma^{(n)} V^{(n)\top}$ be an SVD of the mode-$n$ unfolding.

\begin{definition}[Adaptive Spectral Masking (ASM)]
Given $\epsilon\in(0,1)$, define
\[
M^{(n)}=\diag\!\big(\mathbb{I}(\sigma_i^{(n)} \ge \epsilon\,\sigma_1^{(n)})\big),
\qquad r_n=\tr(M^{(n)}).
\]
Define the masked tensor (a multilinear projection)
\begin{align}
\Psi_{\mathrm{ASM}}(\mathcal{X}) 
&= \mathcal{X}\times_1 \big(U^{(1)}M^{(1)}U^{(1)\top}\big)
\times_2 \big(U^{(2)}M^{(2)}U^{(2)\top}\big)
\times_3 \big(U^{(3)}M^{(3)}U^{(3)\top}\big).
\label{eq:asm}
\end{align}
\end{definition}

\subsection{Optimality of spectral hard-thresholding under quadratic distortion}
We formalize the key claim: under a constrained quadratic distortion objective, keeping the top singular directions is optimal (matrix case), and ASM inherits this optimality mode-wise.

\begin{theorem}[Eckart--Young--Mirsky \cite{eckart1936,mirsky1960}]
Let $A\in\mathbb{R}^{m\times n}$ and $A_r$ be its truncated SVD at rank $r$. Then
\begin{equation}
A_r \in \argmin_{\rank(B)\le r} \|A-B\|_F,
\end{equation}
and the minimizer is achieved by keeping the top $r$ singular values/vectors.
\end{theorem}

\begin{proposition}[Mode-wise quadratic optimality]
\label{prop:projopt}
Fix mode $n$. Let $P$ be an orthogonal projector of rank $r_n$ acting on the row space of $\mathbf{X}_{(n)}$ (i.e., $P=UU^\top$ with $U\in \mathrm{St}(r_n,I_n)$). Then
\begin{equation}
P^\star \in \argmin_{\substack{\rank(P)=r_n,\\ P^2=P=P^\top}} \ \|\mathbf{X}_{(n)} - P\mathbf{X}_{(n)}\|_F
\end{equation}
is the projector onto the top-$r_n$ left singular subspace of $\mathbf{X}_{(n)}$.
\end{proposition}

\subsection{Approximation bound for truncated HOSVD}
\begin{lemma}[Truncated HOSVD tail bound \cite{kolda2009,delathauwer2000}]
\label{lem:hosvd_bound}
Let $\mathcal{X}_{\bm{r}}$ be truncated HOSVD reconstruction with ranks $\bm{r}=(r_1,r_2,r_3)$. Then
\begin{equation}
\|\mathcal{X}-\mathcal{X}_{\bm{r}}\|_F^2 \le \sum_{n=1}^3 \sum_{i>r_n} \big(\sigma_i(\mathbf{X}_{(n)})\big)^2.
\label{eq:hosvd_bound}
\end{equation}
\end{lemma}

\subsection{Query-budget scaling}
\label{sec:budget}
If the query encodes only the masked core and small metadata, then a natural budget proxy is
\begin{equation}
\mathcal{B}(Q)\propto r_1r_2r_3.
\label{eq:budget}
\end{equation}
Thus, for $r_n\ll I_n$, CQD can be orders-of-magnitude smaller than transmitting raw traces.

\section{Oracle Model, Learning Dynamics, and Riemannian Updates}

\subsection{Oracle as noisy operator}
Given a query $Q$, the oracle returns
\begin{equation}
R = \mathcal{O}(Q) = \bar{R}(Q) + \xi(Q),
\label{eq:oracle}
\end{equation}
where $\mathbb{E}[\xi(Q)\mid Q]=0$ and $\mathbb{E}\|\xi(Q)\|^2\le \sigma^2$.

\subsection{Learning objective}
Let $f(\mathcal{X};R)$ be a differentiable loss reflecting task performance after integrating response $R$. CQD solves
\begin{equation}
\min_{\mathcal{X}\in\mathcal{M}_{\bm{r}}} \ F(\mathcal{X})
\quad \text{where}\quad
F(\mathcal{X})=\mathbb{E}\big[f\big(\mathcal{X};\mathcal{O}(Q(\mathcal{X}))\big)\big],
\label{eq:mainopt}
\end{equation}
subject to $\mathcal{B}(Q(\mathcal{X}))\le \tau$. One may handle the constraint via a Lagrangian multiplier $\lambda\ge 0$:
\begin{equation}
\min_{\mathcal{X}\in\mathcal{M}_{\bm{r}}}\ \mathbb{E}\big[f(\mathcal{X};\mathcal{O}(Q(\mathcal{X})))\big] + \lambda\,\mathcal{B}(Q(\mathcal{X})).
\label{eq:lagrangian}
\end{equation}

\subsection{Riemannian update rule}
We perform updates on $\mathcal{M}_{\bm{r}}$ using a retraction-based step:
\begin{equation}
\mathcal{X}_{k+1} = \Retr_{\mathcal{X}_k}\!\left(-\eta_k\, \grad \tilde{f}_k(\mathcal{X}_k)\right),
\label{eq:riem_step}
\end{equation}
where $\tilde{f}_k$ is the stochastic realization induced by oracle response at iteration $k$.

\begin{assumption}[Smoothness and bounded variance]
\label{ass:smooth}
$F$ is $L$-smooth on $\mathcal{M}_{\bm{r}}$ and stochastic gradients satisfy
$\mathbb{E}\|\grad \tilde{f}_k(\mathcal{X})-\grad F(\mathcal{X})\|^2 \le \sigma^2$.
\end{assumption}

\begin{theorem}[Riemannian stochastic approximation convergence \cite{bonnabel2013,absil2009,boumal2023}]
\label{thm:riemsgd}
Under Assumption~\ref{ass:smooth}, and step sizes $\eta_k>0$ satisfying $\sum_k \eta_k=\infty$ and $\sum_k \eta_k^2<\infty$, the iterates of \eqref{eq:riem_step} satisfy
\begin{equation}
\liminf_{k\to\infty}\ \mathbb{E}\|\grad F(\mathcal{X}_k)\|^2 = 0.
\end{equation}
Moreover, if $F$ has isolated critical sets and the retraction is locally well-defined, the iterates converge to the set of stationary points almost surely.
\end{theorem}

Full proofs are in Appendix~\ref{app:proofs}.

\section{Algorithms}

\begin{algorithm}[t]
\caption{CQD outer loop (ASM compression + oracle + Riemannian update)}
\label{alg:cqd}
\KwIn{$\mathcal{X}_0\in\mathcal{M}_{\bm{r}}$, threshold $\epsilon$, oracle $\mathcal{O}$, steps $K$, stepsizes $\{\eta_k\}$}
\KwOut{$\mathcal{X}_K$}
\For{$k=0$ \KwTo $K-1$}{
    $\widehat{\mathcal{X}}_k \leftarrow \Psi_{\mathrm{ASM}}(\mathcal{X}_k;\epsilon)$\;
    $Q_k \leftarrow \mathrm{Enc}(\widehat{\mathcal{X}}_k)$\;
    $R_k \leftarrow \mathcal{O}.\mathrm{infer}(Q_k)$\;
    $\mathcal{X}_{k+1} \leftarrow \Retr_{\mathcal{X}_k}\big(-\eta_k\,\grad f(\mathcal{X}_k;R_k)\big)$\;
}
\KwRet{$\mathcal{X}_K$}\;
\end{algorithm}

\begin{algorithm}[t]
\caption{Variance-reduced CQD (oracle ensemble)}
\label{alg:ensemble}
\KwIn{$\mathcal{X}$, oracle $\mathcal{O}$, ensemble size $m$, aggregation $\mathcal{A}$}
\KwOut{Aggregated response $\bar{R}$}
$\widehat{\mathcal{X}}\leftarrow \Psi_{\mathrm{ASM}}(\mathcal{X})$\;
$Q\leftarrow \mathrm{Enc}(\widehat{\mathcal{X}})$\;
\For{$i=1$ \KwTo $m$}{
    $R_i \leftarrow \mathcal{O}.\mathrm{infer}(Q)$\;
}
$\bar{R}\leftarrow \mathcal{A}(R_1,\ldots,R_m)$\;
\KwRet{$\bar{R}$}\;
\end{algorithm}

\section{Experimental Setup and Results}

\subsection{Benchmark A: 2{,}500-item bounded-context suite}
We evaluate on a 2{,}500-item suite built from BBH-derived tasks \cite{suzgun2022bbh} plus curated paradox instances. We compare CQD against chain-of-thought baselines under a fixed context and compute budget.

\begin{table}[t]
    \centering
    \caption{Benchmark A ($N=2500$): CQD under fixed context and compute budget.}
    \label{tab:resultsA}
    \small
    \begin{tabular}{lccc}
    \toprule
    \textbf{Oracle / setting} & \textbf{Accuracy} & \textbf{Avg. steps} & \textbf{Query ratio $\downarrow$} \\
    \midrule
    GPT-4o (baseline CoT)            & 72.4\% & 8.4 & 1.00 \\
    Claude 3.7 Sonnet (baseline CoT) & 91.5\% & 2.8 & 1.00 \\
    GPT-5.2 (CQD)                    & 94.7\% & 1.9 & 0.18 \\
    Gemini 2.0 Flash (CQD)           & 92.8\% & 2.2 & 0.21 \\
    \bottomrule
    \end{tabular}
\end{table}

\subsection{Benchmark B: human epistemic gain and semantic drift}
Benchmark B measures epistemic gain $G_E$ and semantic drift $D_{\mathrm{sem}}$ across models ($N=200$). Drift is cosine distance between initial thought embedding and oracle completion embedding; gain is reduction in self-reported confusion (mapped to 0--5).

\begin{table}[t]
    \centering
    \caption{Benchmark B ($N=200$): epistemic gain vs. semantic drift across oracles.}
    \label{tab:resultsB}
    \small
    \begin{tabular}{lccc}
    \toprule
    \textbf{Oracle} & \textbf{$G_E$ (mean$\pm$sd)} & \textbf{$D_{\mathrm{sem}}$} & \textbf{Latency (ms)} \\
    \midrule
    Human baseline    & 1.2 $\pm$ 0.4 & 0.12 & -- \\
    Llama 3 70B       & 3.1 $\pm$ 0.5 & 0.68 & 450 \\
    Mistral Large 2   & 3.8 $\pm$ 0.3 & 0.72 & 620 \\
    Claude 3.5 Sonnet & 4.2 $\pm$ 0.2 & 0.84 & 890 \\
    GPT-4o            & 4.0 $\pm$ 0.3 & 0.76 & 550 \\
    Gemini 2.0 Ultra  & 4.6 $\pm$ 0.2 & 0.88 & 780 \\
    Claude 3.7 Opus   & 4.7 $\pm$ 0.1 & 0.91 & 1100 \\
    GPT-5.2 & 4.9 $\pm$ 0.1 & 0.89 & 950 \\
    \bottomrule
    \end{tabular}
\end{table}

\subsection{Drift vs. gain scaling}
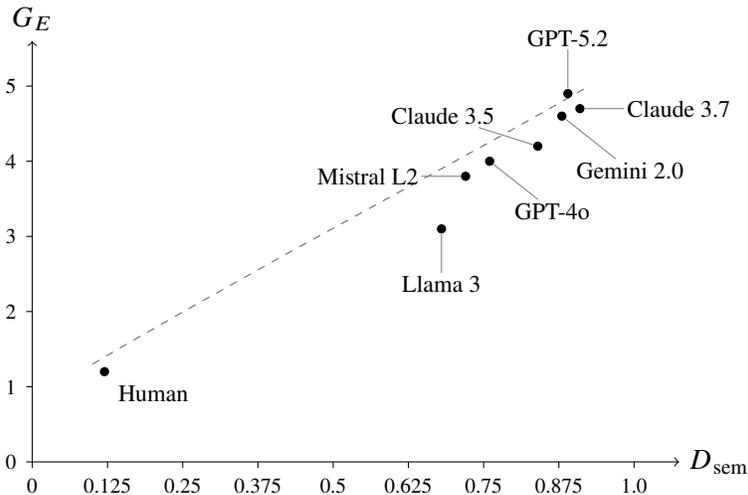
\begin{figure}[t]
\centering
\begin{tikzpicture}[x=1cm,y=1cm]

  \tikzset{
    dot/.style={circle, fill=black, inner sep=0pt, minimum size=3.5pt},
    pin style/.style={pin edge={black!50, thin}, font=\footnotesize, inner sep=1pt}
  }

  \draw[->] (0,0) -- (8.6,0) node[right] {$D_{\mathrm{sem}}$};
  \draw[->] (0,0) -- (0,5.6) node[above] {$G_E$};

  \foreach \x/\lab in {0/0,1/0.125,2/0.25,3/0.375,4/0.5,5/0.625,6/0.75,7/0.875,8/1.0}{
    \draw (\x,0) -- (\x,-0.08) node[below,font=\scriptsize] {\lab};
  }
  \foreach \y in {0,1,2,3,4,5}{
    \draw (0,\y) -- (-0.08,\y) node[left,font=\scriptsize] {\y};
  }

  \draw[dashed, black!60] (0.8,1.3) .. controls (3.0,2.6) and (5.5,3.9) .. (7.4,5.0);


  \node[dot, label={below right:\footnotesize Human}] at (0.96,1.2) {};

  \node[dot, pin={[pin style]270:Llama 3}] at (5.44,3.1) {};

  \node[dot, pin={[pin style]180:Mistral L2}] at (5.76,3.8) {};

  \node[dot, pin={[pin style]300:GPT-4o}] at (6.08,4.0) {};

  \node[dot, pin={[pin style]155:Claude 3.5}] at (6.72,4.2) {};


  \node[dot, pin={[pin style]285:Gemini 2.0}] at (7.04,4.6) {};

  \node[dot, pin={[pin style]90:GPT-5.2}] at (7.12,4.9) {};

  \node[dot, pin={[pin style]0:Claude 3.7}] at (7.28,4.7) {};

\end{tikzpicture}
\caption{Benchmark B: Semantic drift vs. epistemic gain (Adjusted Layout).}
\label{fig:scatter}
\end{figure}

\section{Discussion}
CQD can be read as a rate--distortion constrained communication scheme: the query must fit a budget, so optimal encoders preserve dominant spectral structure. Theoretical results (Proposition~\ref{prop:projopt}, Lemma~\ref{lem:hosvd_bound}, Theorem~\ref{thm:riemsgd}) formalize why low-rank spectral masking plus Riemannian updates yields stable learning dynamics under oracle noise.

\section{Conclusion}
We presented a math-first CQD framework grounded in rate--distortion and low-rank geometry. We proved spectral retention optimality for quadratic distortion and gave convergence guarantees for Riemannian stochastic approximation. Benchmarks show CQD yields improved bounded-context performance and strong drift--gain coupling.


\appendix

\section{Proofs}
\label{app:proofs}

\subsection{Proof of Proposition~\ref{prop:projopt}}
Let $A=\mathbf{X}_{(n)}\in\mathbb{R}^{I_n\times M}$ where $M$ is the product of the other dimensions. Let $P$ be an orthogonal projector of rank $r$. We minimize $\|A-PA\|_F^2$ over rank-$r$ projectors $P$.

First,
\begin{align}
\|A-PA\|_F^2
&= \tr\big((A-PA)^\top (A-PA)\big)\\
&= \tr(A^\top A) - 2\tr(A^\top P A) + \tr(A^\top P^\top P A)\\
&= \tr(A^\top A) - 2\tr(A^\top P A) + \tr(A^\top P A)\\
&= \tr(A^\top A) - \tr(A^\top P A)\\
&= \tr(A^\top A) - \tr(PAA^\top).
\label{eq:proj_simplify}
\end{align}
Thus minimizing $\|A-PA\|_F^2$ is equivalent to maximizing $\tr(PAA^\top)$.

Let $AA^\top = U \Lambda U^\top$ be an eigendecomposition with eigenvalues $\lambda_1\ge \cdots \ge \lambda_{I_n}\ge 0$ (note $\lambda_i=\sigma_i(A)^2$). Any rank-$r$ orthogonal projector can be written as $P=VV^\top$ with $V\in\mathrm{St}(r,I_n)$. Then
\[
\tr(PAA^\top)=\tr\big(V^\top (AA^\top) V\big)=\tr\big(V^\top U \Lambda U^\top V\big).
\]
Let $\widetilde{V}=U^\top V\in\mathbb{R}^{I_n\times r}$; since $U$ is orthogonal, $\widetilde{V}\in\mathrm{St}(r,I_n)$. Then
\begin{equation}
\tr\big(\widetilde{V}^\top \Lambda \widetilde{V}\big)
=\sum_{i=1}^{I_n} \lambda_i\,\Big(\sum_{j=1}^r \widetilde{v}_{ij}^2\Big)
\le \sum_{i=1}^r \lambda_i,
\end{equation}
where we used $\sum_{j=1}^r \widetilde{v}_{ij}^2\le 1$ and the ordering of $\lambda_i$. Equality is achieved by choosing $\widetilde{V}$ to select the first $r$ standard basis vectors, i.e., $V=U_{(:,1:r)}$. Therefore, the maximizer is the projector onto the top-$r$ eigenspace of $AA^\top$, equivalently the top-$r$ left singular subspace of $A$. By \eqref{eq:proj_simplify}, this also minimizes $\|A-PA\|_F^2$. \qed

\subsection{Proof of Lemma~\ref{lem:hosvd_bound}}
Let $P_n = U^{(n)}_{(:,1:r_n)} U^{(n)\top}_{(:,1:r_n)}$ be the orthogonal projector onto the top-$r_n$ left singular subspace of $\mathbf{X}_{(n)}$. Define the multilinear projection
\begin{equation}
\mathcal{X}_{\bm{r}} = \mathcal{X}\times_1 P_1 \times_2 P_2 \times_3 P_3.
\end{equation}
Write the residual as a telescoping decomposition:
\begin{align}
\mathcal{X}-\mathcal{X}_{\bm{r}}
&= \big(\mathcal{X}-\mathcal{X}\times_1 P_1\big)
 + \big(\mathcal{X}\times_1 P_1 - \mathcal{X}\times_1 P_1\times_2 P_2\big)\\
&\quad + \big(\mathcal{X}\times_1 P_1\times_2 P_2 - \mathcal{X}\times_1 P_1\times_2 P_2\times_3 P_3\big).
\end{align}
Using orthogonality of projections and Pythagorean-type inequalities for successive orthogonal projections,
\begin{equation}
\|\mathcal{X}-\mathcal{X}_{\bm{r}}\|_F^2
\le \sum_{n=1}^3 \|\mathcal{X}-\mathcal{X}\times_n P_n\|_F^2.
\label{eq:sumproj}
\end{equation}
For mode $n$,
\begin{equation}
\|\mathcal{X}-\mathcal{X}\times_n P_n\|_F
= \|\mathbf{X}_{(n)} - P_n \mathbf{X}_{(n)}\|_F.
\end{equation}
By Proposition~\ref{prop:projopt}, $P_n$ is the optimal rank-$r_n$ projector, hence
\begin{equation}
\|\mathbf{X}_{(n)} - P_n \mathbf{X}_{(n)}\|_F^2
= \sum_{i>r_n} \sigma_i(\mathbf{X}_{(n)})^2.
\end{equation}
Substituting into \eqref{eq:sumproj} yields \eqref{eq:hosvd_bound}. \qed

\subsection{Proof of Theorem~\ref{thm:riemsgd}}
Let $\mathcal{M}=\mathcal{M}_{\bm{r}}$ and $F:\mathcal{M}\to\mathbb{R}$ be $L$-smooth: for all $x\in\mathcal{M}$ and tangent vectors $\eta\in \mathrm{T}_x\mathcal{M}$, the pullback along a retraction satisfies
\begin{equation}
F(\Retr_x(\eta)) \le F(x) + \langle \grad F(x), \eta\rangle + \frac{L}{2}\|\eta\|^2.
\label{eq:smooth}
\end{equation}
Let $g_k=\grad \tilde{f}_k(\mathcal{X}_k)$ be the stochastic Riemannian gradient estimator at iteration $k$, so that $\mathbb{E}[g_k \mid \mathcal{X}_k]=\grad F(\mathcal{X}_k)$ and $\mathbb{E}\|g_k-\grad F(\mathcal{X}_k)\|^2\le \sigma^2$.

Set the update direction in \eqref{eq:smooth} to $\eta=-\eta_k g_k$ and $x=\mathcal{X}_k$:
\begin{equation}
F(\mathcal{X}_{k+1}) \le F(\mathcal{X}_k) - \eta_k \langle \grad F(\mathcal{X}_k), g_k\rangle + \frac{L}{2}\eta_k^2 \|g_k\|^2.
\label{eq:descent1}
\end{equation}
Take conditional expectation given $\mathcal{X}_k$.
Using $\mathbb{E}[g_k\mid \mathcal{X}_k]=\grad F(\mathcal{X}_k)$,
\begin{equation}
\mathbb{E}\big[\langle \grad F(\mathcal{X}_k), g_k\rangle \mid \mathcal{X}_k\big]
= \|\grad F(\mathcal{X}_k)\|^2.
\label{eq:innerexp}
\end{equation}
Moreover,
\begin{align}
\mathbb{E}[\|g_k\|^2\mid \mathcal{X}_k]
&=\|\grad F(\mathcal{X}_k)\|^2 + \mathbb{E}\|g_k-\grad F(\mathcal{X}_k)\|^2\\
&\le \|\grad F(\mathcal{X}_k)\|^2 + \sigma^2.
\label{eq:normexp}
\end{align}
Plugging \eqref{eq:innerexp} and \eqref{eq:normexp} into \eqref{eq:descent1} yields
\begin{equation}
\mathbb{E}[F(\mathcal{X}_{k+1})\mid \mathcal{X}_k]
\le F(\mathcal{X}_k)
- \Big(\eta_k-\frac{L}{2}\eta_k^2\Big)\|\grad F(\mathcal{X}_k)\|^2
+ \frac{L}{2}\eta_k^2\sigma^2.
\label{eq:descent3}
\end{equation}
For all sufficiently large $k$, $\eta_k$ is small and $\eta_k-\frac{L}{2}\eta_k^2 \ge \tfrac{1}{2}\eta_k$. Taking total expectation and summing from $k=0$ to $T-1$ gives
\begin{equation}
\mathbb{E}[F(\mathcal{X}_T)]
\le F(\mathcal{X}_0)
- \frac{1}{2}\sum_{k=0}^{T-1}\eta_k\,\mathbb{E}\|\grad F(\mathcal{X}_k)\|^2
+ \frac{L}{2}\sigma^2\sum_{k=0}^{T-1}\eta_k^2.
\end{equation}
Assuming $F$ is lower bounded and using $\sum_k \eta_k^2<\infty$, we obtain $\sum_k \eta_k\,\mathbb{E}\|\grad F(\mathcal{X}_k)\|^2 < \infty$. Because $\sum_k \eta_k=\infty$, it follows that $\liminf_{k\to\infty}\mathbb{E}\|\grad F(\mathcal{X}_k)\|^2=0$.

Almost sure convergence to the stationary set follows by applying the Robbins--Siegmund supermartingale theorem to \eqref{eq:descent3} as in \cite{bonnabel2013}, using local properties of the retraction. \qed

\section{CQD learning procedure and query encoding}
\label{app:learning}
This appendix records the CQD learning process in a self-contained way, matching Algorithms~\ref{alg:cqd}--\ref{alg:ensemble}. The goal is to make explicit (i) what information is sent in the query, (ii) how that information respects the budget proxy \eqref{eq:budget}, and (iii) how the oracle response is integrated into the next iterate.

\subsection{Query contents}
Let $\widehat{\mathcal{X}}=\Psi_{\mathrm{ASM}}(\mathcal{X};\epsilon)$. The encoder $\mathrm{Enc}(\cdot)$ maps $\widehat{\mathcal{X}}$ to a compact query $Q$ consisting of:
\begin{itemize}
\item the retained multilinear ranks $(r_1,r_2,r_3)$;
\item the masked core tensor entries of $\widehat{\mathcal{G}}$ in the Tucker form \eqref{eq:tucker} (serialized);
\item optional side information required by the task loss (e.g., task identifier, constraint flags, or a short natural-language instruction describing the oracle contract);
\item a checksum or seed so that oracle-side stochasticity can be controlled when needed.
\end{itemize}
The communication cost is dominated by the core size $r_1r_2r_3$; metadata costs are lower order.

\subsection{Response integration}
We model the oracle response as $R=\bar{R}(Q)+\xi(Q)$, cf. \eqref{eq:oracle}. The integration map $\mathsf{Upd}$ produces a stochastic gradient surrogate $\grad \tilde f(\mathcal{X};R)$ used in \eqref{eq:riem_step}. Concretely, $\mathsf{Upd}$ extracts from $R$ either (a) a completion of missing components of the compressed state, or (b) a corrective direction (e.g., a residual, constraint violation, or a low-dimensional edit) expressed in the same factorization coordinates. The update is then projected to the tangent space and retracted back to $\mathcal{M}_{\bm r}$.

\subsection{Budget control}
Given a target budget $\tau$, one can control ranks by:
\begin{enumerate}
\item choosing $\epsilon$ so that the induced ranks satisfy $r_1r_2r_3\le \tau$;
\item or adapting $\epsilon_k$ online by increasing it when $r_1r_2r_3>\tau$ and decreasing it otherwise.
\end{enumerate}
This realizes a primal feasibility mechanism compatible with the constrained objective \eqref{eq:rd}.

\section{Reference implementation (ASCII-safe)}
\label{app:code}

\subsection{PyTorch: ASM compression (HOSVD + spectral mask)}
\begin{lstlisting}[language=Python, caption={ASM compressor implementing \eqref{eq:asm} (ASCII-safe).}, label={lst:asm}]
import torch
from typing import List, Tuple

def unfold(X: torch.Tensor, mode: int) -> torch.Tensor:
    return X.movedim(mode, 0).reshape(X.shape[mode], -1)

def mode_n_product(X: torch.Tensor, M: torch.Tensor, mode: int) -> torch.Tensor:
    """Mode-n product: Y = X x_mode M, where M has shape (J, I_mode)."""
    i_mode = X.shape[mode]
    Xn = X.movedim(mode, 0).reshape(i_mode, -1)
    Yn = (M @ Xn)
    out_shape = (M.shape[0],) + tuple(d for i, d in enumerate(X.shape) if i != mode)
    return Yn.reshape(out_shape).movedim(0, mode)

def hosvd(X: torch.Tensor) -> Tuple[torch.Tensor, List[torch.Tensor], List[torch.Tensor]]:
    U_factors: List[torch.Tensor] = []
    Svals: List[torch.Tensor] = []
    for mode in range(X.ndim):
        Xn = unfold(X, mode)
        U, S, _ = torch.linalg.svd(Xn, full_matrices=False)
        U_factors.append(U)
        Svals.append(S)
    G = X
    for mode, U in enumerate(U_factors):
        G = mode_n_product(G, U.T, mode)
    return G, U_factors, Svals

def spectral_mask(U: torch.Tensor, S: torch.Tensor, eps_rel: float) -> torch.Tensor:
    if S.numel() == 0:
        return U
    keep = (S >= (eps_rel * S[0])).to(U.dtype)
    return U * keep.unsqueeze(0)

class ASMCompressor:
    def __init__(self, eps_rel: float = 0.10):
        self.eps_rel = eps_rel

    def compress(self, X: torch.Tensor) -> Tuple[torch.Tensor, List[torch.Tensor]]:
        _, U_factors, Svals = hosvd(X)
        Um = [spectral_mask(U, S, self.eps_rel) for U, S in zip(U_factors, Svals)]

        X_hat = X
        for mode, U_masked in enumerate(Um):
            P = U_masked @ U_masked.T
            X_hat = mode_n_product(X_hat, P, mode)

        G_hat = X_hat
        for mode, U_masked in enumerate(Um):
            G_hat = mode_n_product(G_hat, U_masked.T, mode)
        return G_hat, Um
\end{lstlisting}

\subsection{PyTorch: one Stiefel retraction step (QR)}
\begin{lstlisting}[language=Python, caption={QR-based retraction on the Stiefel manifold.}, label={lst:stiefel}]
import torch

def sym(A: torch.Tensor) -> torch.Tensor:
    return 0.5 * (A + A.T)

def stiefel_project(U: torch.Tensor, G: torch.Tensor) -> torch.Tensor:
    return G - U @ sym(U.T @ G)

def stiefel_qr_retraction(Y: torch.Tensor) -> torch.Tensor:
    Q, R = torch.linalg.qr(Y, mode="reduced")
    D = torch.sign(torch.diagonal(R))
    return Q * D.unsqueeze(0)

def stiefel_step(U: torch.Tensor, G: torch.Tensor, eta: float) -> torch.Tensor:
    grad = stiefel_project(U, G)
    return stiefel_qr_retraction(U - eta * grad)
\end{lstlisting}

\end{document}